\documentclass[10pt,twocolumn,letterpaper]{article}

\usepackage{cvpr}
\usepackage{times}
\usepackage{epsfig}
\usepackage{graphicx}
\usepackage{amsmath}
\usepackage{amssymb}

\usepackage{color}
\usepackage{array}
\usepackage{multirow}
\usepackage{subfig}
\usepackage{epstopdf}

\usepackage[pagebackref=true,breaklinks=true,letterpaper=true,colorlinks,bookmarks=false]{hyperref}

\cvprfinalcopy 

\def\cvprPaperID{3437} 

\ifcvprfinal\fi
\begin{document}

\title{Image Super-Resolution by Neural Texture Transfer}

\author{Zhifei Zhang\\
Adobe Research\\
{\tt\small zzhang@adobe.com}
\and
Zhaowen Wang\\
Adobe Research\\
{\tt\small zhawang@adobe.com}
\and
Zhe Lin\\
Adobe Research\\
{\tt\small zlin@adobe.com}
\and
Hairong Qi\\
University of Tennessee\\
{\tt\small hqi@utk.edu}
}

\newcolumntype{L}[1]{>{\raggedright\let\newline\\\arraybackslash\hspace{0pt}}m{#1}}
\newcolumntype{C}[1]{>{\centering\let\newline\\\arraybackslash\hspace{0pt}}m{#1}}
\newcolumntype{R}[1]{>{\raggedleft\let\newline\\\arraybackslash\hspace{0pt}}m{#1}}
\newcommand{\alg}{SRNTT}
\newcommand{\dataset}{CUFED5}
\newcommand{\shrink}{\vspace{-15pt}}
\newcommand{\zw}[1]{\textcolor{red}{[ZW] #1}}

\maketitle

\begin{abstract}
Due to the significant information loss in low-resolution (LR) images, it has become extremely challenging to further advance the state-of-the-art of single image super-resolu-tion (SISR). Reference-based super-resolution (RefSR), on the other hand,  has proven to be promising in recovering high-resolution (HR) details when a reference (Ref) image with similar content as that of the LR input is given. However, the quality of RefSR can degrade severely when Ref is less similar. This paper aims to unleash the potential of RefSR by leveraging more texture details from Ref images with stronger robustness even when irrelevant Ref images are provided.
Inspired by the recent work on image stylization, we formulate the RefSR problem as neural texture transfer. We design an end-to-end deep model which enriches HR details by adaptively transferring the texture from Ref images according to their textural similarity. Instead of matching content in the raw pixel space as done by previous methods, our key contribution is a multi-level matching conducted in the neural space. This matching scheme facilitates multi-scale neural transfer that allows the model to benefit more from those semantically related Ref patches, and gracefully degrade to SISR performance on the least relevant Ref inputs. We build a benchmark dataset for the general research of RefSR, which contains Ref images paired with LR inputs with varying levels of similarity. Both quantitative and qualitative evaluations demonstrate the superiority of our method over state-of-the-art\footnote{Code: \url{https://github.com/ZZUTK/SRNTT}}.

\end{abstract}

\section{Introduction}


The traditional single image super-resolution (SISR) problem is defined as recovering a high-resolution (HR) image from its low-resolution (LR) observation~\cite{yang2014single}. 
As in other fields of computer vision studies, the introduction of convolutional neural networks (CNNs)~\cite{dong2014learning,wang2015deep,kim2016deeply,lim2017enhanced,wang2018recovering,han2018image} has greatly advanced the state-of-the-art of SISR. However, due to the ill-posed nature of SISR problems, most existing methods still suffer from blurry results at large upscaling factors, \eg, 4$\times$, especially when it comes to the recovery of fine texture present in the original HR image but lost in its LR counterpart. 
In recent years, perceptual-related constraints, \eg, perception loss~\cite{johnson2016perceptual} and adversarial loss~\cite{goodfellow2014generative}, have been introduced to the SISR problem formulation, leading to major breakthroughs on visual quality under large upscaling factors~\cite{ledig2017photo,sajjadi2017enhancenet}. However, they tend to hallucinate fake textures and even produce artifacts.
\newcommand{\wdemo}{.24\columnwidth}
\begin{figure*}[t]
	\centering
	\includegraphics[width=2\columnwidth]{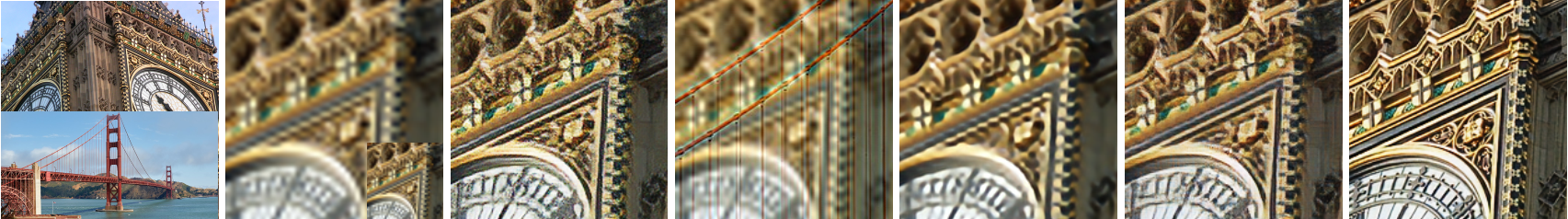}
	\begin{tabular}{C{\wdemo}C{\wdemo}C{\wdemo}C{\wdemo}C{\wdemo}C{\wdemo}C{\wdemo}}
		Ref images & Bicubic \& LR & SRGAN & CrossNet (L) & CrossNet (U) & SRNTT (L) & SRNTT (U)
	\end{tabular}
	\caption{
    \alg~(ours) is compared to SRGAN~\cite{ledig2017photo} (a state-of-the-art SISR method) and CrossNet~\cite{zheng2018crossnet} (a state-of-the-art RefSR method). (a) Two Ref images. The upper one (U) has similar content to the LR input as shown in (b) bottom-right corner, and the lower one (L) has distinct or unrelated content to the LR input. (c) Result of SRGAN. (d)(e) Results of CrossNet using two Ref images respectively. (f)(g) Results of  \alg~using two Ref images respectively.
    }
	\label{fig:demo}
\end{figure*}

This paper diverts from the traditional SISR and explores the reference-based super-resolution (RefSR). RefSR utilizes rich textures from the HR references (Ref) to compensate for the lost details in the LR images, relaxing the ill-posed issue and producing more detailed and realistic textures with the help of reference images. Note that the Ref images can be obtained from various sources like photo albums, video frames, web image search, etc.
There are existing RefSR approaches~\cite{freeman2002example,chang2004super,freedman2011image,sun2012super,yue2013landmark,timofte2013anchored,liu2017retrieval,zheng2018crossnet} that adopt internal examples (self-example) or external high-frequency information to enhance textures. However, these approaches assume the reference images possess similar content as that of the LR image and/or with good alignment. Otherwise, their performance would significantly degrade and even become worse than SISR methods.
In contrast, the Ref images play a different role in our setting: it does not require well alignment or similar content to the LR image. Instead, we only intend to transfer the semantically relevant texture from Ref images to the output SR image.
Ideally, a robust RefSR algorithm should outperform SISR when good Ref images are given, and achieve comparable performance as SISR when Ref images are not provided or do not possess relevant texture at all. Note that content similarity would infer texture similarity but not vice versa. 

Inspired by the recent work on image stylization~\cite{gatys2016image,johnson2016perceptual,chen2016fast}, we propose a new RefSR algorithm, named Super-Resolution by Neural Texture Transfer (SRNTT), which \textit{adaptively} transfers textures from the Ref images to the SR image.
More specifically, \alg~conducts local texture matching in the feature space and transfers matched textures to the final output through a deep model. The texture transfer model learns the complicated dependency between LR and Ref textures, and leverages similar textures while suppressing dissimilar textures.
The example in Fig.~\ref{fig:demo} illustrates the advantage of the proposed \alg~compared with two state-of-the-art works, \ie, SRGAN~\cite{ledig2017photo} (for SISR) and CrossNet~\cite{zheng2018crossnet} (for RefSR). \alg~shows significant boost in synthesizing finer texture as compared to the other methods if using a Ref image with similar content (\ie, Fig.~\ref{fig:demo}(a) upper). Even using a Ref image with unrelated content (\ie, Fig.~\ref{fig:demo}(a) lower), \alg~is still comparable to SRGAN (similar visual quality but less artifacts), demonstrating the adaptiveness/robustness of \alg~to different Ref images of various levels of content similarity.
By contrast, CrossNet would introduce undesired textures from the unrelated Ref image and shows severe performance degradation


In order to facilitate fair comparison and help advance research on the RefSR problem in general, we propose a new dataset, named \dataset, which provides training and testing sets accompanied with references of different similarity levels in terms of content, texture, color, illumination, view point, etc. 
The main contributions of this paper are:
\begin{itemize}
	\item We explore a more general RefSR problem, breaking the performance barrier in SISR (\ie, lack of texture detail) and relaxing constraints in existing RefSR (\ie, alignment assumption). 
	\item We propose an end-to-end deep model, \alg, for the RefSR problem to recover the LR image conditioned on any given references by multi-scale neural texture transfer. 
	We demonstrate the visual improvement, effectiveness, and adaptiveness of the proposed \alg~by extensive empirical studies.   
	\item We build a benchmark dataset,  \dataset, to facilitate the further research and performance evaluation of RefSR methods in handling references with different levels of similarity to the LR input image.  
	
\end{itemize}

In the rest of this paper, we review the related works in Section~\ref{sec:related_work}. The network architecture and training criteria are discussed in Section~\ref{sec:approach}. In Section~\ref{sec:dataset}, the proposed dataset \dataset~is described in detail. The results of both quantitative and qualitative evaluations are presented in Section~\ref{sec:experiment}. Finally, Section~\ref{sec:conclusion} concludes this paper.

\section{Related Works}
\label{sec:related_work}

\subsection{Deep Learning based SISR}

In recent years, deep learning based SISR has shown superior performance in terms of either PSNR or visual quality compared to non-deep-learning based methods~\cite{dong2014learning,wang2015deep,ledig2017photo}. The reader could refer to~\cite{nasrollahi2014super,yang2014single} for more comprehensive review. Here we will only focus on deep learning based methods.

A milestone work that introduced CNN into SR was proposed by Dong et al.~\cite{dong2014learning}, where a three-layer fully convolutional network was trained to minimize the mean squared error (MSE) between the SR image and the original HR image. It demonstrated the effectiveness of deep learning in SR and achieved the state-of-the-art performance. Wang et al.~\cite{wang2015deep} combined the strengths of sparse coding and deep network and made considerable improvement over previous models. To speed up the SR process, Dong et al.~\cite{dong2016accelerating} and Shi et al.~\cite{shi2016real} extracted features directly from the LR image, that also achieved better performance compared to processing the upscaled LR image through bicubic interpolation. 
In recent years, the state-of-the-art performance (in PSNR) were all achieved by deep learning based models~\cite{kim2016deeply,kim2016accurate,lim2017enhanced}.

The above mentioned methods, in general, aim at minimizing MSE between the SR and HR images, which might not always be consistent with the human evaluation (\ie, perceptual quality)~\cite{ledig2017photo,sajjadi2017enhancenet}. 
Therefore, perceptual-related constraints were incorporated to achieve better visual quality. Johnson et al.~\cite{johnson2016perceptual} demonstrated the effectiveness of adding perception loss using VGG~\cite{simonyan2014very}. Ledig et al.~\cite{ledig2017photo} introduced adversarial loss from the generative adversarial nets (GANs)~\cite{goodfellow2014generative} to minimize the perceptually relevant distance between the SR and HR images. Sajjadi et al.~\cite{sajjadi2017enhancenet} further incorporated the texture matching loss based on the idea of style transfer~\cite{gatys2015texture,gatys2016image} to enhance the texture in the SR image. The proposed \alg~is more closely related to \cite{ledig2017photo,sajjadi2017enhancenet}, where perceptual-related constraints (\ie, perceptual loss and adversarial loss) are incorporated to recover more visually plausible SR images.

\subsection{Reference-based Super-Resolution}

In contrast to SISR where only a single LR image is used as input, RefSR methods introduce additional images to assist the SR process. In general, the reference images need to possess similar texture and/or content structure with the LR image. The references could be selected from adjacent frames in a video~\cite{liu2011bayesian,caballero2017real}, images from web retrieval~\cite{yue2013landmark}, an external database (dictionary)~\cite{zhu2014single}, or images from different view points~\cite{zheng2018crossnet}. 
There is a batch of SR methods that refer to self patches/neighborhood~\cite{freeman2002example,chang2004super,freedman2011image,Huang-CVPR-2015}, which are widely known as self-example based SR. They do not utilize external references, thus more close to SISR problems.
These works mostly build the mapping from LR to HR patches and fuse the HR patches at the pixel level or by a shallow model, which is insufficient to model the complicated dependency between the LR image and extracted details from the HR patches. 
A more generic scenario of utilizing the references was proposed by Yue et al.~\cite{yue2013landmark}, which instantly retrieves similar images from web and conducts global registration and local matching. However, they made a strong assumption --- the references have to be well aligned to the LR image. In addition, the shallow model for patch blending made its performance highly dependent on how well the references could be aligned. 
Zheng et al.~\cite{zheng2018crossnet} proposed a deep model based RefSR method and adopted optical flow to align input and reference. However, optical flow is limited in matching long distance correspondences, thus incapable of handling significantly misaligned references.
The proposed \alg~adopts the ideas of local texture (patch) matching which could handle long distance dependency. Like existing RefSR methods, we also ``fuse'' Ref texture to the final output, but we conduct it in the multi-scale feature space through a deep model, which enables the learning of complicated transfer process from references with scaling, rotation, or even non-rigid deformations. 

\section{Approach}
\label{sec:approach}

\begin{figure}[t]
	\centering
	\includegraphics[width=.9\columnwidth]{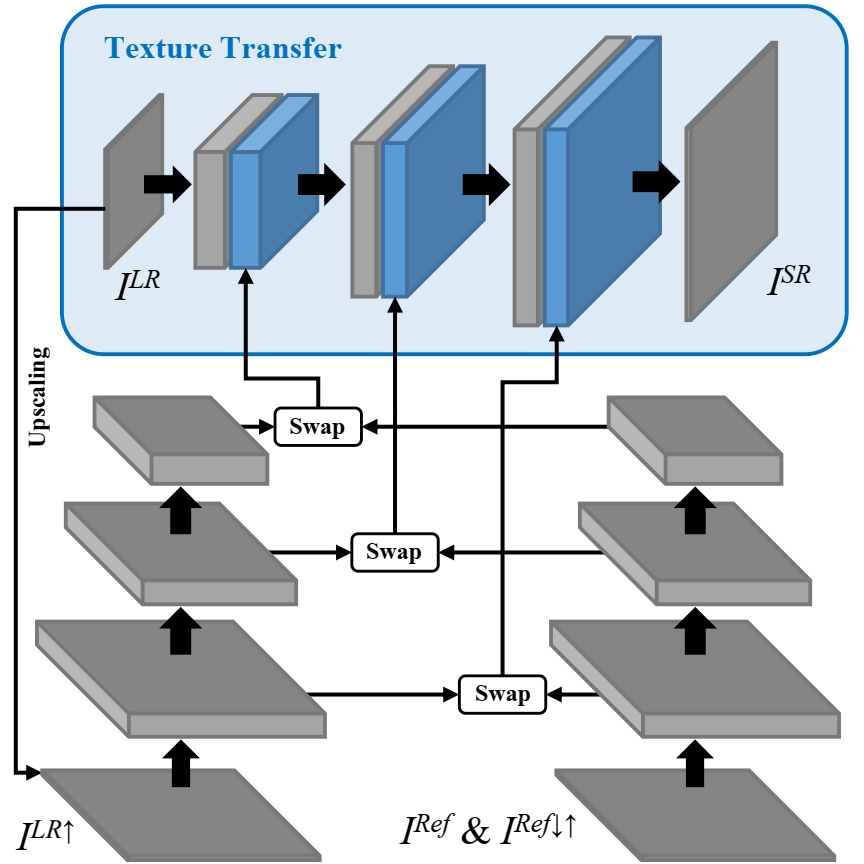}
	\caption{The proposed \alg~framework with feature swapping and texture transfer.
	} 
	\label{fig:flow}
\end{figure}

The proposed \alg~aims to estimate the SR image $I^{SR}$ from its LR counterpart $I^{LR}$ and the given reference images $I^{Ref}$, synthesizing plausible textures conditioned on $I^{Ref}$ while preserving the consistency with $I^{LR}$ in content. 
An overview of the proposed \alg~is shown in Fig.~\ref{fig:flow}.
The main idea is to search for matching texture from $I^{Ref}$ in the feature space and then transfer matched textures to $I^{SR}$ in a multi-scale fashion, since the features are more robust to the variance of color and illumination. The multi-scale texture transfer simultaneously considers semantic (higher-level) and textual (lower-level) similarity between $I^{LR}$ and $I^{Ref}$, leading to transferring related textures while suppressing irrelevant textures.

In addition to minimizing the pixel and/or perceptual distance between the output $I^{SR}$ and the original HR image $I^{HR}$ as most existing SR methods do, we further regularize on the texture consistency between $I^{SR}$ and the matched textures from $I^{Ref}$, enforcing the effectiveness of texture transfer. The final output $I^{SR}$ is synthesized in an end-to-end manner. Texture searching and transfer will be discussed in Sections~\ref{subsec:swap} and \ref{subsec:transfer}, respectively. Section~\ref{subsec:losses} will detail the objective function of \alg.

\subsection{Feature Swapping}  
\label{subsec:swap}
We first conduct feature swapping which searches over the entire $I^{Ref}$ for locally similar textures that can be used to replace (or swap) the texture features of $I^{LR}$ for enhanced SR recovery.
The feature searching is conducted in HR spatial coordinate to enable direct texture transfer to the final output $I^{SR}$. 
Following the self-example matching strategy \cite{freedman2011image}, 
we first apply bicubic up-sampling on $I^{LR}$ to get an upscaled LR image $I^{LR\uparrow}$ that has the same spatial size as $I^{HR}$. 
We also sequentially apply bicubic down-sampling and up-sampling with the same factor on $I^{Ref}$ to obtain a blurry Ref image $I^{Ref\downarrow\uparrow}$ that matches the frequency band of $I^{LR\uparrow}$.
Instead of estimating a global transformation or optical flow, we match the local patches in $I^{LR\uparrow}$ and $I^{Ref\downarrow\uparrow}$ so that there is no constraint on the global structure of the Ref image, which is a key advantage over CrossNet~\cite{zheng2018crossnet}.
As LR and Ref patches may also differ in color and illumination, we match their similarity in the neural feature space $\phi(I)$ to emphasize the structural and textural information.
We use inner product to measure the similarity between neural features:
\begin{equation}
s_{i,j} = \left< P_i(\phi(I^{LR\uparrow})), \frac{P_j(\phi(I^{Ref\downarrow\uparrow}))}{\|P_j(\phi(I^{Ref\downarrow\uparrow}))\|} \right>  ,
\label{eq:inner}
\end{equation} 
where $P_i(\cdot)$ denotes sampling the $i$-th patch from neural feature map, and $s_{i,j}$ is the similarity between the $i$-th LR patch and the $j$-th Ref patch. The Ref patch feature is normalized for selecting the best match over all $j$.
The similarity computation can be efficiently implemented as a set of convolution (or correlation) operations over all LR patches with each kernel corresponding to a Ref patch:
\begin{equation}
S_j = \phi(I^{LR\uparrow}) \ast \frac{P_j(\phi(I^{Ref\downarrow\uparrow}))}{\|P_j(\phi(I^{Ref\downarrow\uparrow}))\|} ,
\label{eq:conv}
\end{equation}
where $S_j$ is the similarity map for the $j$-th Ref patch, and $\ast$ denotes the correlation operation. We use $S_j(x, y)$ to denote the similarity between the LR patch centered at location $(x, y)$ and the $j$-th Ref patch. Both LR and Ref patches are densely sampled from their images.
Based on the similarity score, we can construct a swapped feature map $M$ to represent texture-enhanced LR image. Each patch in $M$ centered at $(x, y)$ is defined as
\begin{equation}
P_{\omega(x,y)}(M) = P_{j^*}(\phi(I^{Ref})), \; j^*{=}\arg\underset{j}{\max} S_j(x,y) ,
\label{eq:swap}
\end{equation}
where $\omega(\cdot,\cdot)$ maps patch center to patch index.
Note that while $I^{Ref\downarrow\uparrow}$ is used for matching (Eq.~\ref{eq:conv}), the raw Ref $I^{Ref}$ is used in swapping (Eq.~\ref{eq:swap}) so that the HR information from the original references is preserved.
Due to the dense sampling of LR patches, we take the average of the swapped features $P_{j^*}(\phi(I^{Ref}))$ in the regions where they overlap.
The resulting swapped feature map $M$ is used as the basis for the next texture transfer stage.


\subsection{Neural Texture Transfer}  
\label{subsec:transfer}

Our texture transfer model is designed by merging multiple swapped texture feature maps into a base deep generative network at different feature layers corresponding to various scales, as illustrated in Fig.~\ref{fig:flow} (blue box). For each scale or neural layer $l$, a swapped feature map $M_l$ is constructed using the method introduced above, with a texture feature encoder $\phi_l$ matching the current scale.
The effectiveness of transferring texture across multiple layers is verified by the ablation study in Section~\ref{subsec:ablation}. 

\begin{figure}[t]
	\centering
	\includegraphics[width=.9\columnwidth]{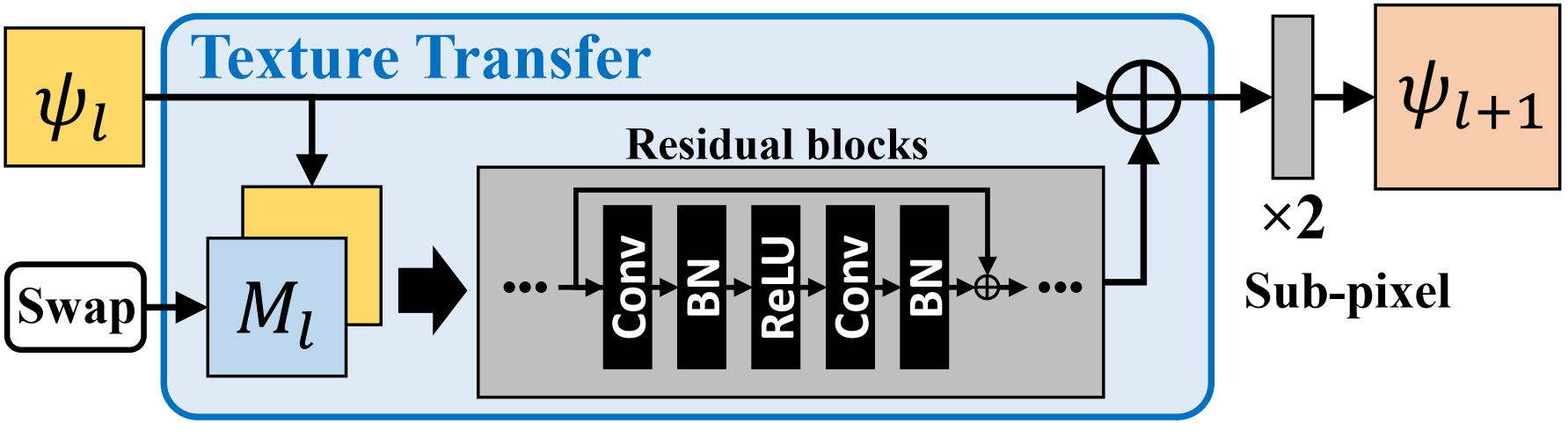}
	\caption{The network structure for texture transfer.}
	\label{fig:transfer}
\end{figure}

We use residual blocks and skip connections~\cite{he2016deep,he2016identity,ledig2017photo} to build the base generative network.
The network output $\psi_l$ at layer $l$ is defined recursively as
\begin{equation}
\psi_l = \left[ \text{Res} \left( \psi_{l-1} \| M_{l-1} \right)  + \psi_{l-1} \right]\uparrow_{2\times},
\label{eq:transfer}
\end{equation}
where $\text{Res}(\cdot)$ denotes the residual blocks, $\|$ denotes channel-wise concatenation, and $\uparrow_{2\times}$ denotes $2\times$ upscaling with sub-pixel convolution~\cite{shi2016real}.
The final SR result image is generated after $L$ layers to reach target HR resolution:
\begin{equation}
I^{SR} =  \text{Res} \left( \psi_{L-1} \| M_{L-1} \right)  + \psi_{L-1} 
\label{eq:transfer2}
\end{equation}
Fig.~\ref{fig:transfer} illustrates the network structure of texture transfer at one scale, where the residual blocks extract related texture from $M_l$ (\ie, $I^{Ref}$) conditioned on $\psi_l$ (\ie, $I^{LR}$) and merge it with target content.

Different from traditional SISR methods that only reduce the difference between $I^{SR}$ and the ground truth $I^{HR}$, our proposed \alg~method further takes into account the texture difference between $I^{SR}$ and $I^{Ref}$.
That is, we require the texture of $I^{SR}$ to be similar as the swapped feature map $M_l$ in the feature space of $\phi_l$.
Specifically, we define a texture loss $\mathcal{L}_{tex}$ as
\begin{equation}
\mathcal{L}_{tex}{=}\sum_l \lambda_l \left\| Gr \left( \phi_l(I^{SR}) \cdot S^*_l \right) - Gr\left( M_l  \cdot S^*_l  \right) \right\|_F,
\label{eq:l_tex}
\end{equation}
where $Gr(\cdot)$ computes the Gram matrix, and $\lambda_l$ is a normalization factor corresponding to the feature size of layer $l$.
$S^*_l$ is a weighting map for all LR patches calculated as the best matching score in Eq.~\ref{eq:swap}. Intuitively, textures dissimilar to $I^{LR}$ will have lower weight, and thus receiving lower penalty in texture transfer.
In this way, the texture transfer from $I^{Ref}$ to $I^{SR}$ is adaptively enforced based on the Ref image quality, leading to more robust texture hallucination as demonstrated in Section~\ref{subsec:ablation}.

\subsection{Training Objective}
\label{subsec:losses}
In order to 1) preserve the spatial structure of the LR image, 2) improve the visual quality of the SR image, and 3) take advantage of the rich texture from Ref images, our objective function combines reconstruction loss $\mathcal{L}_{rec}$, perceptual loss $\mathcal{L}_{per}$, adversarial loss $\mathcal{L}_{adv}$, and texture loss $\mathcal{L}_{tex}$. 
The reconstruction loss is adopted in most SR methods. The perceptual and adversarial losses improve visual quality. The texture loss already discussed in Eq.~\ref{eq:l_tex} is specific to RefSR.

\textbf{Reconstruction loss} aims to achieve higher PSNR, usually measured in terms of mean square error (MSE). In this paper, we adopt the $\ell_1$-norm,
\begin{equation}
\mathcal{L}_{rec} = \left\|I^{HR}-I^{SR}\right\|_1,
\label{eq:l_rec}
\end{equation} 
The $\ell_1$-norm would further sharpen $I^{SR}$ as compared to MSE. In addition, it is consistent to the objective of WGAN-GP, which will be discussed later in the adversarial loss.

\textbf{Perceptual loss} has been investigated in recent SR works~\cite{bruna2016super,johnson2016perceptual,ledig2017photo,sajjadi2017enhancenet} for better visual quality. We adopt the relu5\_1 layer of VGG19~\cite{simonyan2014very}, 
\begin{equation}
\mathcal{L}_{per}=\frac{1}{V}\sum_{i=1}^{C}\left\|\phi_i(I^{HR})-\phi_i(I^{SR})\right\|_F,
\end{equation}
where $V$ and $C$ indicate the volume and channel number of the feature maps, respectively, and $\phi_i$ denotes the $i$th channel of the feature maps extracted from the hidden layer of VGG19 model. $\|\cdot\|_F$ denotes the Frobenius norm.   

\textbf{Adversarial loss} could significantly enhance the sharpness/visual quality of synthesized images~\cite{isola2017image,zhang2017age}. 
Here, we adopt WGAN-GP~\cite{gulrajani2017improved}, which improves upon WGAN by penalizing the gradient, achieving more stable results. Because the Wasserstein distance in WGAN is based on $\ell_1$-norm, we use $\ell_1$-norm as the reconstruction loss (Eq.~\ref{eq:l_rec}). 
Intuitively, consistent objectives would facilitate the optimization process. 
The adversarial loss is expressed as
\begin{align}
\mathcal{L}_{adv} = &  -\mathbb{E}_{\tilde{x}\sim\mathbb{P}_g}[D(\tilde{x})], 
\\
\underset{G}{\min}\;\underset{D\in\mathcal{D}}{\max} \; \mathbb{E}_{x\sim\mathbb{P}_r}[D(x)] & -\mathbb{E}_{\tilde{x}\sim\mathbb{P}_g}[D(\tilde{x})], 
\end{align}
where $\mathcal{D}$ is the set of 1-Lipschitz functions, and $\mathbb{P}_r$ and $\mathbb{P}_g$ are the model distribution and real distribution, respectively.   


\subsection{Implementation Details}

We adopt a pre-trained VGG19~\cite{simonyan2014very} model for feature swapping, which is well-known for its power of texture representation~\cite{gatys2015texture,gatys2016image}. Feature layers relu1\_1, relu2\_1, and relu3\_1 are used as texture encoder $\phi_l$'s in multiple scales. To speed up the matching process, we only match on the relu3\_1 layer and project the correspondence to layers relu2\_1 and relu1\_1, and use the same correspondence across all layers.
The weights for $\mathcal{L}_{rec}$, $\mathcal{L}_{per}$, $\mathcal{L}_{adv}$, and $\mathcal{L}_{tex}$ are 1, 1e-4, 1e-6, and 1e-4, respectively. Adam optimizer is used with the learning rate of 1e-4. The network is pre-trained for 2 epochs, where only $\mathcal{L}_{rec}$ is applied. Then, all losses are involved to train another 20 epochs.

Our method can be easily extended to handle multiple Ref images. In all our RefSR experiments, we augment each $I^{Ref}$ with its scaled and rotated versions to get more accurate texture matching results.


\section{Dataset}
\label{sec:dataset}
For RefSR problems, the similarity between the LR and Ref images affects SR results significantly. In general, references with various levels of similarity to LR images should be provided for the purpose of both training and evaluating a RefSR algorithm.
To the best of our knowledge, there has not been such a dataset available for public usage. 
We thus construct such a dataset with Ref images at various similarity levels based on the CUFED~\cite{Wang_16_CVPR} dataset that contains 1,883 albums capturing diverse events in daily life. The size of each album varies between 30 and 100 images.
Within each album, we collect image pairs in different similarity levels based on SIFT~\cite{lowe1999object} feature matching, which characterizes local texture pattern that is in line with the objective of local texture matching.

\begin{figure}[t]
	\centering
	\includegraphics[width=\columnwidth]{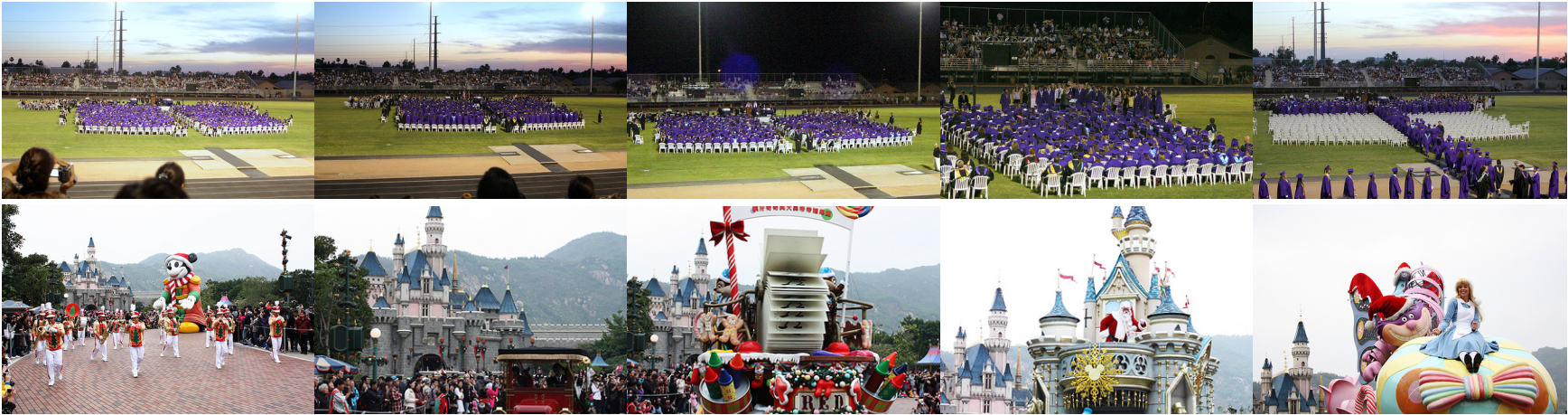}
	
	\caption{Examples from the \dataset~testing set. From left to right are HR image and the corresponding Ref images of similarity levels L1, L2, L3 and L4, respectively.}
	\label{fig:test_set}
\end{figure}

We define four similarity levels from high to low, \ie, L1, L2, L3, and L4, according to the number of best matches of SIFT features. From each paired images, we randomly crop 160$\times$160 patches from one image as the original HR images, and the corresponding references are cropped from the other image. In this way, we collect 13,761 paired patches as the training set.
For the testing dataset, each HR image is paired with all four levels of references in order to extensively evaluate the adaptiveness of a reference-based SR method.
We use the similar way to collect image pairs as in building the training dataset. In total, the testing set contains 126 groups of samples. Each group consists of one HR image and four references at levels L1, L2, L3, and L4, respectively. Two examples from the testing set are shown in Fig.~\ref{fig:test_set}. 
We refer to the collected training and testing sets as \dataset, which would largely facilitate the research on RefSR and provide a benchmark for fair comparison.

To evaluate the generalization capacity of the trained model on \dataset, we test it on Sun80~\cite{sun2012super} and Urban100~\cite{Huang-CVPR-2015}. The Sun80 dataset has 80 natural images, each of which is accompanied by a series of web-searching references, while the Urban100 dataset contains building images without references. 

\section{Experimental Results}
\label{sec:experiment}
In this section, both quantitative and qualitative comparisons are conducted to demonstrate the advantages of the proposed \alg~in terms of visual quality and texture enrichment.
Following standard protocol, we obtain all LR images by bicubic downscaling (4$\times$) from the HR images.


\subsection{Quantitative Evaluation}
We compare the proposed \alg~with the state-of-the-art SISR and RefSR algorithms\footnote{
	Implementation of SR algorithms in comparison:\\ 
	\tiny
	SRCNN: \url{http://mmlab.ie.cuhk.edu.hk/projects/SRCNN.html} \\
	SelfEx: \url{https://sites.google.com/site/jbhuang0604/publications/struct_sr} \\
	SCN: \url{http://www.ifp.illinois.edu/~dingliu2/iccv15/} \\
	DRCN: \url{http://cv.snu.ac.kr/research/DRCN/} \\
	LapSRN: \url{http://vllab.ucmerced.edu/wlai24/LapSRN/} \\
	MDSR: \url{https://github.com/LimBee/NTIRE2017} \\
	ENet: \url{https://webdav.tue.mpg.de/pixel/enhancenet/} \\
	SRGAN: \url{https://github.com/tensorlayer/srgan} \\
	CrossNet: \url{https://github.com/htzheng/ECCV2018_CrossNet_RefSR}
} as shown in Table~\ref{tab:cmp}. The SISR methods in comparison are SRCNN~\cite{dong2014learning}, SelfEx~\cite{Huang-CVPR-2015}, SCN~\cite{wang2015deep}, DRCN~\cite{kim2016deeply}, LapSRN~\cite{lai2017deep}, MDSR~\cite{lim2017enhanced}, ENet~\cite{sajjadi2017enhancenet}, and SRGAN~\cite{ledig2017photo}, among which  MDSR~\cite{lim2017enhanced} has achieved the state-of-the-art performance in PSNR in recent two years, while ENet~\cite{sajjadi2017enhancenet} and SRGAN~\cite{ledig2017photo} are considered the state-of-the-art in visual quality.
Two RefSR methods are also included in the comparison, \ie, Landmark~\cite{yue2013landmark} and the recently proposed CrossNet~\cite{zheng2018crossnet}, which outperforms previous RefSR methods.

\begin{table}[t]
	\centering
	\caption{
		PSNR/SSIM comparison of different SR methods on three datasets. Methods are grouped by SISR (top) and RefSR (bottom) with their respective best numbers in bold.
	}
	\label{tab:cmp}
	\footnotesize
	\begin{tabular}{l|ccc}
		\hline
		\centering Algorithm & CUFED5 & Sun80 \cite{sun2012super} & Urban100 \cite{huang2015single}  \\
		\hline
		Bicubic & 24.18 / 0.684 & 27.24 / 0.739 & 23.14 / 0.674 \\
		SRCNN~\cite{dong2014learning} &  25.33 / 0.745 & 28.26 / 0.781 & 24.41 / 0.738 \\
		SelfEx~\cite{Huang-CVPR-2015} & 23.22 / 0.680 & 27.03 / 0.756 & 24.67 / 0.749 \\
		SCN~\cite{wang2015deep} & 25.45 / 0.743 & 27.93 / 0.786  & 24.52 / 0.741 \\
		DRCN~\cite{kim2016deeply} & 25.26 / 0.734 & 27.84 / 0.785 & 25.14 / 0.760\\
		LapSRN~\cite{lai2017deep} & 24.92 / 0.730 & 27.70 / 0.783  & 24.26 / 0.735 \\
		MDSR~\cite{lim2017enhanced} & \textbf{25.93} / \textbf{0.777} & \textbf{28.52} / \textbf{0.792}  & \textbf{25.51} / \textbf{0.783} \\
		ENet~\cite{sajjadi2017enhancenet} & 24.24 / 0.695 & 26.24 / 0.702 & 23.63 / 0.711 \\
		SRGAN~\cite{ledig2017photo} & 24.40 / 0.702 & 26.76 / 0.725 & 24.07 / 0.729 \\
        SRNTT-$\ell_2$ (SISR)&  25.91 / 0.776 &  28.46 / 0.790 &   25.50 / 0.783 \\
        \hline
		Landmark~\cite{yue2013landmark}  & 24.91 / 0.718 & 27.68 / 0.776  & --- \\
		CrossNet~\cite{zheng2018crossnet} & 25.48 / 0.764   & 28.52 / 0.793  & 25.11 / 0.764\\
		\alg-$\ell_2$ & \textbf{26.24} / \textbf{0.784} & \textbf{28.54} / \textbf{0.793}  & \textbf{25.50} / \textbf{0.783} \\
		SRNTT &  25.61 / 0.764 &  27.59 / 0.756 &   25.09 / 0.774 \\
		\hline
	\end{tabular}
\end{table}

\newcommand{\wh}{.40\columnwidth}
\newcommand{\ww}{.45\columnwidth}
\newcommand{\hcut}{90}
\begin{figure*}[ht]
	\centering
	\begin{tabular}{C{\wh}C{\wh}C{\wh}C{\wh}}
		Truth & MDSR~\cite{lim2017enhanced} & ENet~\cite{sajjadi2017enhancenet} & \alg-$\ell_2$ (ours) \\
		\hline
		Reference & CrossNet~\cite{zheng2018crossnet} & SRGAN~\cite{ledig2017photo} & \alg~(ours)\\
	\end{tabular}
	\includegraphics[width=1.84\columnwidth]{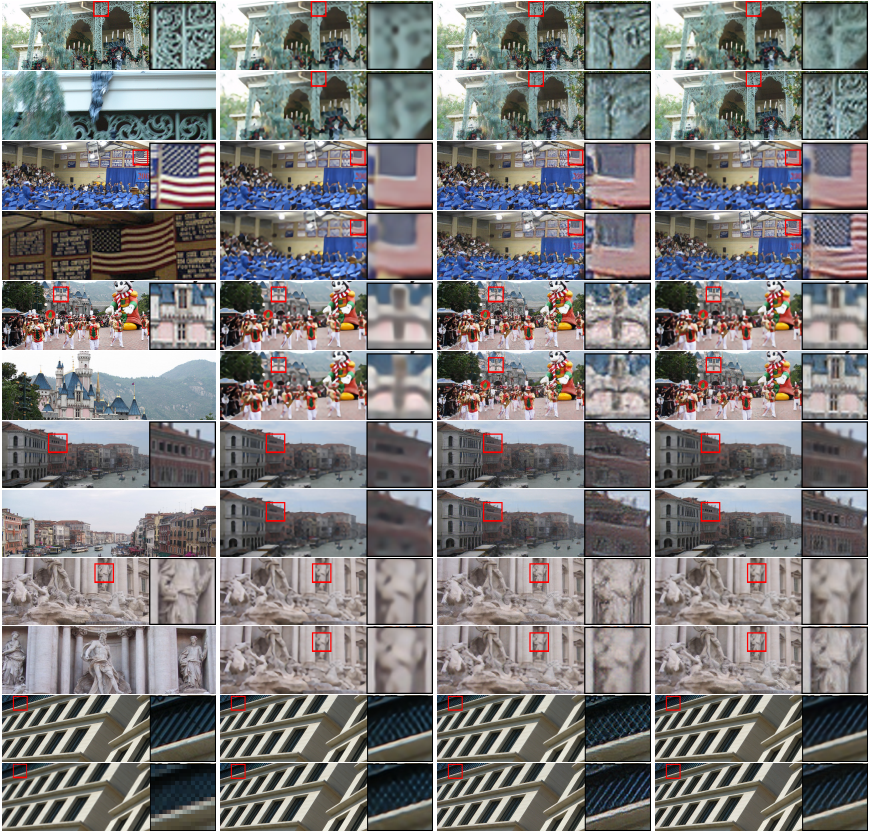}
	\caption{Visual comparison among different SR methods on \dataset~(top three examples), Sun80~\cite{sun2012super} (the forth and fifth examples), and Urban100~\cite{Huang-CVPR-2015} (the bottom example whose reference image is the LR input).}
	\label{fig:cmp}
\end{figure*}

For fair comparison, all learning-based methods are trained on the proposed \dataset~dataset, and tested on \dataset, Sun80~\cite{sun2012super}, and Urban100~\cite{Huang-CVPR-2015}, respectively. For fair comparison on PSNR/SSIM with those methods mainly minimizing MSE, \eg, SCN and MDSR, we train a simplified version of \alg~by only minimizing the MSE, \ie, \alg-$\ell_2$. 
Note that Table~\ref{tab:cmp} shows the results of \alg-$\ell_2$ in both SISR (upper block) and RefSR (lower block) settings. Specifically, the \alg-$\ell_2$ under SISR setting uses the LR input as reference. 
In \dataset~and Sun80 datasets, each input corresponds to multiple references, all of which are used in Landmark, \alg-$\ell_2$ and \alg, while CrossNet uses the reference that yields the highest PSNR because CrossNet accepts only one reference. 

In Table~\ref{tab:cmp}, \alg-$\ell_2$~achieves the highest score on \dataset~and Sun80 which have references, while performing comparably to MDSR (the highest score) on Urban100 which does not have references. Even with SISR setting on all datasets, \alg-$\ell_2$ (SISR) performs similarly to the state-of-the-art. The proposed \alg, which uses adversarial loss that would increase visual quality but reduce PSNR, outperforms ENet and SRGAN in PSNR (even comparable to those methods that only minimize MSE), while at the same time achieving higher visual quality (finer texture and less artifacts) as shown by the examples in Fig.~\ref{fig:cmp}. A more comprehensive evaluation on visual quality will be conducted in Section~\ref{subsec:survey}. As demonstrated by the examples, \alg~outperforms CrossNet in recovering fine texture from references. The main reason is that the references present large disparity/misalignment from the LR image, which CrossNet is incapable of handling. 

Without loss of generality, examples from Sun80 and Urban100 are displayed in Fig.~\ref{fig:cmp}. With the help of references, \alg~outperforms other SR methods on Sun80. On Urban100, however, there is no HR references. We use LR input as the reference and achieve finer texture that could be transferred from the LR image. In general, \alg~would outperform existing SR methods with the assistance of references, and we could still achieve state-of-the-art SISR performance when there is no HR information from references. Section~\ref{subsec:ablation} will further demonstrate the adaptiveness of \alg~by analyzing the performance on references of different similarity levels.

\subsection{Qualitative Evaluation by User Study}
\label{subsec:survey}

\begin{figure}[t]
	\centering
	\includegraphics[width=.78\columnwidth]{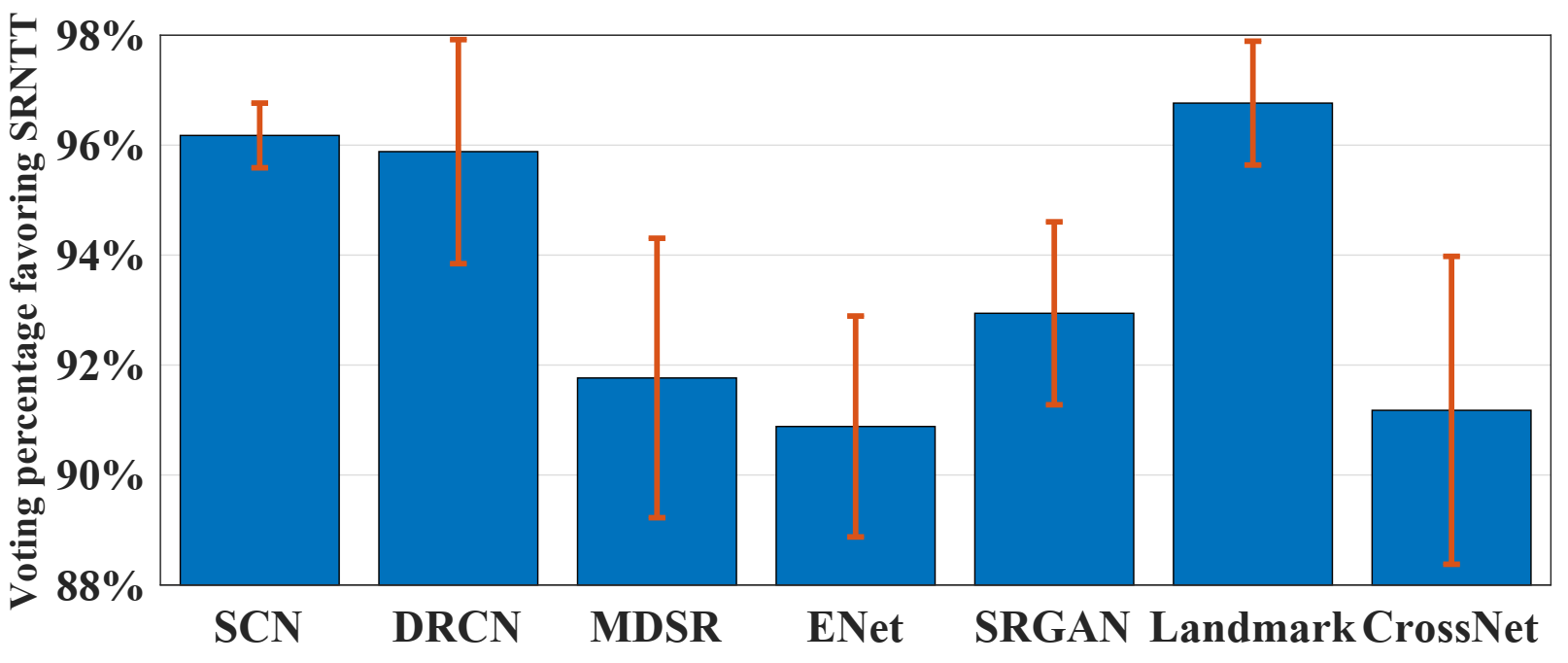}
	\caption{The user study result. \alg~is compared to each algorithm along the horizontal axis, and the blue bars indicate the percentage of users favoring \alg~results. 
    }
	\label{fig:votes_bar}
\end{figure}

To evaluate the visual quality of the SR images, we conduct user study, where \alg~ is compared to SCN~\cite{wang2015deep}, DRCN~\cite{kim2016deeply}, MDSR~\cite{lim2017enhanced}, ENet~\cite{sajjadi2017enhancenet}, SRGAN~\cite{ledig2017photo}, Landmark~\cite{yue2013landmark}, and CrossNet~\cite{zheng2018crossnet}. 
We present the users with pair-wise comparisons, \ie, \alg~vs.~other, and ask the users to select the one with higher resolution. For each reference level, 2,400 votes are collected on the testing results from the \dataset~dataset. 
Fig.~\ref{fig:votes_bar} shows the voting results,  
where the percentages favoring \alg~denotes the percentage of users that prefer \alg~as compared to the algorithms denoted along the horizontal axis. 
Overall, \alg~significantly outperforms the other algorithms with over 90\% users voting for \alg. 

\subsection{Ablation Studies} 
\label{subsec:ablation}

\subsubsection{Effect of reference similarity}
Similarity between LR and Ref images is a key factor to the success of RefSR methods. This section investigates the performance of CrossNet~\cite{zheng2018crossnet} and the proposed \alg~ at different reference levels. Table~\ref{tab:cmp_levels} lists the results at six levels of references, where ``HR (warp)'' denotes the reference obtained by random translation (quarter to half width/height), rotation (10$\sim$30 degree), and scaling (1.2$\times$$\sim$2.0$\times$ upscaling) from the original HR image. L1, L2, L3, and L4 are the four levels of references from the proposed \dataset~dataset. ``LR'' means using the LR input image as the references (there is no external references). 
\begin{table*}[ht]
	\centering
	\caption{
		PSNR/SSIM at different reference levels on \dataset~dataset. PM indicates if patch-based matching is used; GAN indicates if GAN and other perceptual losses are used.}
	\label{tab:cmp_levels}
	\begin{tabular}{l|cc|cccccc}
		\hline
		~  & PM & GAN & HR (warp) & L1 & L2 & L3 & L4 & LR\\
		\hline
		CrossNet~\cite{zheng2018crossnet} &  & & 25.49 / .764  & 25.48 / .764 & 25.48 / .764 & 25.47 / .763 & 25.46 / .763 & 25.46 / .763 \\
        \alg-$\ell_2$ & \checkmark & & 29.29 / .889 & \textbf{26.15} / \textbf{.781} & \textbf{26.04} / \textbf{.776} & \textbf{25.98}  / \textbf{.775} & \textbf{25.95} / \textbf{.774} & \textbf{25.91} / \textbf{.776} \\ 
        \hline
        \alg-flow &  & \checkmark  & 25.82 / .801 & 24.64 / .743 & 24.22 / .723 & 24.15 / .719 & 24.05 / .714 & 25.50 / .756\\
		\alg & \checkmark & \checkmark  &  \textbf{33.87} / \textbf{.959} & 25.42 / .758 & 25.32 / .752 &	25.24 / .751 &	25.23 / .750 & 25.10 / .750 \\ 
		\hline
	\end{tabular}
\end{table*}
As compared to CrossNet, the \alg-$\ell_2$~shows superior results at each reference level. At the ``HR'' level, \alg-$\ell_2$~achieves significant improvement, which demonstrates the advantage of patch-wise matching over the alignment using optical flow. 
Comparing \alg~and \alg-$\ell_2$, \alg~shows even higher PSNR at ``HR'' level but lower at other levels. This phenomenon emphasizes the effectiveness of texture loss in recovering fine textures when given highly similar references.  

To further investigate the gap between the CrossNet and SRNTT, we conduct an experiment by replacing feature swapping with optical flow (FlowNet2~\cite{ilg2017flownet}) in the \alg~framework. As shown in Table~\ref{tab:cmp_levels}, ``\alg-flow'' shows large degradation even at ``HR'' level as compared to \alg, reflecting the limitation of optical flow in handling large disparity/misalignment. As the reference similarity level decreases, PSNR/SSIM of \alg~reduces gracefully as well. At ``LR'' level, \alg~still achieves comparable performance as the state-of-the-art SISR algorithms (Table~\ref{tab:cmp}). 
We observe that the PSNR of \alg-flow is higher than that of \alg~at the ``LR'' level because the Ref is identical to the LR input. In this case, optical flow would easily align Ref to LR, while patch matching may have missed some matches. 



\vspace{-6pt}
\subsubsection{Layers for feature swapping}
As discussed in Section~\ref{sec:approach}, feature swapping and transfer at multiple scales would increase the performance of \alg. Table~\ref{tab:ablation} demonstrates the effectiveness of utilizing multiple scales as compared to using single scale. The relu1/2/3 denotes three layers/scales, \ie, relu1\_1, relu2\_1, and relu3\_1 from VGG19, used in \alg~for feature swapping. We observe that 
the performance in PSNR  decreases as reducing the number of scales. The relu3 gets the lowest PSNR because relu3\_1 is a higher-level layer that carries less high-frequency information, contributing less to texture transfer as compared to relu1\_1 and relu2\_1. For each reference level, the PSNR follows the similar trend as the number of scales increases. However, it is interesting that relu3 shows decreasing and then increasing trend as the reference similarity decreases. This demonstrates the stronger adaptiveness of relu3 in preserving spacial structure, \ie, low-similarity textures from the references are suppressed, and it tends to focus more on spacial reconstruction instead of textural recovery. Therefore, the multi-scale texture transfer using deep model gains extreme momentum on adaptively learning the complicated transfer process between the content and external texture. 

\begin{table}[t]
	\centering
	\caption{
		PSNR of using different VGG layers for feature swapping on different reference levels.}
    	\label{tab:ablation}
	\begin{tabular}{c|ccccc}
		\hline
		Layer & relu1 & relu2 & relu3 & relu1/2 & relu1/2/3\\
		\hline
        HR & 28.39 & 28.66 & 24.83 &  30.39 & 33.87\\
        L1 & 24.76 & 24.91 & 24.48 &  25.05 & 25.42\\
        L2 & 24.68 & 24.86 & 24.22 &  25.00 & 25.32\\
        L3 & 24.64 & 24.80 & 24.39 &  24.94 & 25.24\\
        L4 & 24.63 & 24.79 & 24.45 &  24.92 & 25.23\\
		\hline
	\end{tabular}
\end{table}

\subsubsection{Effect of texture loss}
\begin{figure}[ht]
\centering
    \includegraphics[width=.45\columnwidth, trim=120 0 0 90, clip]{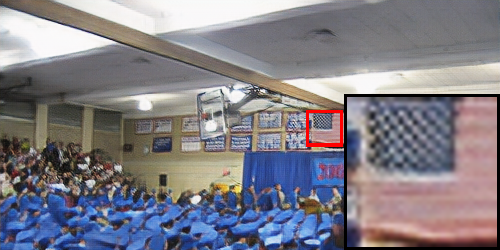}
    \hspace{1 pt}
    \includegraphics[width=.45\columnwidth, trim=120 0 0 90, clip]{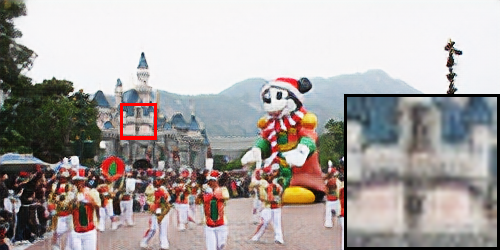}
    \caption{
    SR results with texture loss disabled have degraded quality compared with the same examples in Fig.~\ref{fig:cmp}.
    }
    \label{fig:cmp_tex}
\end{figure}

The weighted texture loss used in the proposed \alg~is a key difference from most SR methods. 
Unlike those style transfer works, where the content image is significantly modified to carry the texture from the style image (\ie, the reference), the proposed \alg~avoids such ``stylization'' by local matching, adaptive neural transfer, and spatial/perceptual regularization. 
The local matching ensures spatially consistent texture, neural transfer gains adaptiveness on texture transfer, and spatial/perceptual regularization forces the spacial consistency globally.   
The effect of texture loss is shown in Fig.~\ref{fig:cmp_tex}. The PSNR tested on \dataset~are 25.25 and 25.61 for \alg~w/o and with the texture loss, respectively. Without the texture loss, the finer texture from the references cannot be effectively transferred into the output.

\vspace{-4 pt}
\section{Conclusion}
\label{sec:conclusion}
This paper exploited the more generic RefSR problem where the references can be arbitrary images. 
We proposed \alg, an end-to-end network structure that performs multi-level adaptive texture transfer from the references to recover more plausible texture in the SR image. Both quantitative and qualitative experiments were conducted to demonstrate the effectiveness and adaptiveness of \alg. In addition, a new dataset \dataset~was constructed to facilitate the evaluation of RefSR methods. It also provides a benchmark for future RefSR research.     

\clearpage
{
	\bibliographystyle{ieee}
	\bibliography{egbib}

\begin{thebibliography}{10}\itemsep=-1pt

\bibitem{bruna2016super}
J.~Bruna, P.~Sprechmann, and Y.~LeCun.
\newblock Super-resolution with deep convolutional sufficient statistics.
\newblock In {\em International Conference on Learning Representations (ICLR)},
  2016.

\bibitem{caballero2017real}
J.~Caballero, C.~Ledig, A.~Aitken, A.~Acosta, J.~Totz, Z.~Wang, and W.~Shi.
\newblock Real-time video super-resolution with spatio-temporal networks and
  motion compensation.
\newblock In {\em IEEE Conference on Computer Vision and Pattern Recognition
  (CVPR)}, 2017.

\bibitem{chang2004super}
H.~Chang, D.-Y. Yeung, and Y.~Xiong.
\newblock Super-resolution through neighbor embedding.
\newblock In {\em IEEE Conference on Computer Vision and Pattern Recognition
  (CVPR)}, 2004.

\bibitem{chen2016fast}
T.~Q. Chen and M.~Schmidt.
\newblock Fast patch-based style transfer of arbitrary style.
\newblock In {\em Workshop in Constructive Machine Learning}. Advances in
  Neural Information Processing Systems, 2016.

\bibitem{dong2014learning}
C.~Dong, C.~C. Loy, K.~He, and X.~Tang.
\newblock Learning a deep convolutional network for image super-resolution.
\newblock In {\em European Conference on Computer Vision (ECCV)}, 2014.

\bibitem{dong2016accelerating}
C.~Dong, C.~C. Loy, and X.~Tang.
\newblock Accelerating the super-resolution convolutional neural network.
\newblock In {\em European Conference on Computer Vision (ECCV)}, 2016.

\bibitem{freedman2011image}
G.~Freedman and R.~Fattal.
\newblock Image and video upscaling from local self-examples.
\newblock {\em ACM Transactions on Graphics}, 30(2):12, 2011.

\bibitem{freeman2002example}
W.~T. Freeman, T.~R. Jones, and E.~C. Pasztor.
\newblock Example-based super-resolution.
\newblock {\em IEEE Computer graphics and Applications}, 22(2):56--65, 2002.

\bibitem{gatys2015texture}
L.~Gatys, A.~S. Ecker, and M.~Bethge.
\newblock Texture synthesis using convolutional neural networks.
\newblock In {\em Advances in Neural Information Processing Systems}, 2015.

\bibitem{gatys2016image}
L.~A. Gatys, A.~S. Ecker, and M.~Bethge.
\newblock Image style transfer using convolutional neural networks.
\newblock In {\em IEEE Conference on Computer Vision and Pattern Recognition
  (CVPR)}, 2016.

\bibitem{goodfellow2014generative}
I.~Goodfellow, J.~Pouget-Abadie, M.~Mirza, B.~Xu, D.~Warde-Farley, S.~Ozair,
  A.~Courville, and Y.~Bengio.
\newblock Generative adversarial nets.
\newblock In {\em Advances in neural information processing systems}, 2014.

\bibitem{gulrajani2017improved}
I.~Gulrajani, F.~Ahmed, M.~Arjovsky, V.~Dumoulin, and A.~C. Courville.
\newblock Improved training of {Wasserstein} {GAN}s.
\newblock In {\em Advances in Neural Information Processing Systems}, 2017.

\bibitem{han2018image}
W.~Han, S.~Chang, D.~Liu, M.~Yu, M.~Witbrock, and T.~S. Huang.
\newblock Image super-resolution via dual-state recurrent networks.
\newblock In {\em IEEE Conference on Computer Vision and Pattern Recognition
  (CVPR)}, 2018.

\bibitem{he2016deep}
K.~He, X.~Zhang, S.~Ren, and J.~Sun.
\newblock Deep residual learning for image recognition.
\newblock In {\em IEEE Conference on Computer Vision and Pattern Recognition
  (CVPR)}, 2016.

\bibitem{he2016identity}
K.~He, X.~Zhang, S.~Ren, and J.~Sun.
\newblock Identity mappings in deep residual networks.
\newblock In {\em European Conference on Computer Vision (ECCV)}, 2016.

\bibitem{Huang-CVPR-2015}
J.-B. Huang, A.~Singh, and N.~Ahuja.
\newblock Single image super-resolution from transformed self-exemplars.
\newblock In {\em IEEE Conference on Computer Vision and Pattern Recognition
  (CVPR)}, 2015.

\bibitem{huang2015single}
J.-B. Huang, A.~Singh, and N.~Ahuja.
\newblock Single image super-resolution from transformed self-exemplars.
\newblock In {\em IEEE Conference on Computer Vision and Pattern Recognition
  (CVPR)}, 2015.

\bibitem{ilg2017flownet}
E.~Ilg, N.~Mayer, T.~Saikia, M.~Keuper, A.~Dosovitskiy, and T.~Brox.
\newblock Flownet 2.0: Evolution of optical flow estimation with deep networks.
\newblock In {\em IEEE Conference on Computer Vision and Pattern Recognition
  (CVPR)}, 2017.

\bibitem{isola2017image}
P.~Isola, J.-Y. Zhu, T.~Zhou, and A.~A. Efros.
\newblock Image-to-image translation with conditional adversarial networks.
\newblock In {\em IEEE conference on computer vision and pattern recognition},
  2017.

\bibitem{johnson2016perceptual}
J.~Johnson, A.~Alahi, and L.~Fei-Fei.
\newblock Perceptual losses for real-time style transfer and super-resolution.
\newblock In {\em European Conference on Computer Vision (ECCV)}, 2016.

\bibitem{kim2016accurate}
J.~Kim, J.~Kwon~Lee, and K.~Mu~Lee.
\newblock Accurate image super-resolution using very deep convolutional
  networks.
\newblock In {\em IEEE Conference on Computer Vision and Pattern Recognition
  (CVPR)}, 2016.

\bibitem{kim2016deeply}
J.~Kim, J.~Kwon~Lee, and K.~Mu~Lee.
\newblock Deeply-recursive convolutional network for image super-resolution.
\newblock In {\em IEEE Conference on Computer Vision and Pattern Recognition
  (CVPR)}, 2016.

\bibitem{lai2017deep}
W.-S. Lai, J.-B. Huang, N.~Ahuja, and M.-H. Yang.
\newblock Deep {Laplacian} pyramid networks for fast and accurate
  super-resolution.
\newblock In {\em IEEE Conference on Computer Vision and Pattern Recognition
  (CVPR)}, 2017.

\bibitem{ledig2017photo}
C.~Ledig, L.~Theis, F.~Huszar, J.~Caballero, A.~Cunningham, A.~Acosta,
  A.~Aitken, A.~Tejani, J.~Totz, Z.~Wang, et~al.
\newblock Photo-realistic single image super-resolution using a generative
  adversarial network.
\newblock In {\em IEEE Conference on Computer Vision and Pattern Recognition
  (CVPR)}, 2017.

\bibitem{lim2017enhanced}
B.~Lim, S.~Son, H.~Kim, S.~Nah, and K.~M. Lee.
\newblock Enhanced deep residual networks for single image super-resolution.
\newblock In {\em The IEEE Conference on Computer Vision and Pattern
  Recognition (CVPR) Workshops}, 2017.

\bibitem{liu2011bayesian}
C.~Liu and D.~Sun.
\newblock A {Bayesian} approach to adaptive video super resolution.
\newblock In {\em IEEE Conference on Computer Vision and Pattern Recognition
  (CVPR)}, 2011.

\bibitem{liu2017retrieval}
J.~Liu, W.~Yang, X.~Zhang, and Z.~Guo.
\newblock Retrieval compensated group structured sparsity for image
  super-resolution.
\newblock {\em IEEE Transactions on Multimedia}, 19(2):302--316, 2017.

\bibitem{lowe1999object}
D.~G. Lowe.
\newblock Object recognition from local scale-invariant features.
\newblock In {\em IEEE International Conference on Computer Vision (ICCV)},
  1999.

\bibitem{nasrollahi2014super}
K.~Nasrollahi and T.~B. Moeslund.
\newblock Super-resolution: a comprehensive survey.
\newblock {\em Machine Vision and Applications}, 25(6):1423--1468, 2014.

\bibitem{sajjadi2017enhancenet}
M.~S. Sajjadi, B.~Scholkopf, and M.~Hirsch.
\newblock {EnhanceNet}: Single image super-resolution through automated texture
  synthesis.
\newblock In {\em IEEE Conference on Computer Vision and Pattern Recognition
  (CVPR)}, 2017.

\bibitem{shi2016real}
W.~Shi, J.~Caballero, F.~Husz{\'a}r, J.~Totz, A.~P. Aitken, R.~Bishop,
  D.~Rueckert, and Z.~Wang.
\newblock Real-time single image and video super-resolution using an efficient
  sub-pixel convolutional neural network.
\newblock In {\em IEEE Conference on Computer Vision and Pattern Recognition
  (CVPR)}, 2016.

\bibitem{simonyan2014very}
K.~Simonyan and A.~Zisserman.
\newblock Very deep convolutional networks for large-scale image recognition.
\newblock In {\em International Conference on Learning Representations (ICLR)},
  2015.

\bibitem{sun2012super}
L.~Sun and J.~Hays.
\newblock Super-resolution from internet-scale scene matching.
\newblock In {\em IEEE International Conference on Computational Photography
  (ICCP)}, 2012.

\bibitem{timofte2013anchored}
R.~Timofte, V.~De, and L.~Van~Gool.
\newblock Anchored neighborhood regression for fast example-based
  super-resolution.
\newblock In {\em IEEE International Conference on Computer Vision (ICCV)},
  2013.

\bibitem{wang2018recovering}
X.~Wang, K.~Yu, C.~Dong, and C.~Change~Loy.
\newblock Recovering realistic texture in image super-resolution by deep
  spatial feature transform.
\newblock In {\em IEEE Conference on Computer Vision and Pattern Recognition
  (CVPR)}, 2018.

\bibitem{Wang_16_CVPR}
Y.~Wang, Z.~Lin, X.~Shen, R.~Mech, G.~Miller, and G.~W. Cottrell.
\newblock Event-specific image importance.
\newblock In {\em IEEE Conference on Computer Vision and Pattern Recognition
  (CVPR)}, 2016.

\bibitem{wang2015deep}
Z.~Wang, D.~Liu, J.~Yang, W.~Han, and T.~Huang.
\newblock Deep networks for image super-resolution with sparse prior.
\newblock In {\em IEEE International Conference on Computer Vision (ICCV)},
  2015.

\bibitem{yang2014single}
C.-Y. Yang, C.~Ma, and M.-H. Yang.
\newblock Single-image super-resolution: A benchmark.
\newblock In {\em European Conference on Computer Vision (ECCV)}, 2014.

\bibitem{yue2013landmark}
H.~Yue, X.~Sun, J.~Yang, and F.~Wu.
\newblock Landmark image super-resolution by retrieving web images.
\newblock {\em IEEE Transactions on Image Processing}, 22(12):4865--4878, 2013.

\bibitem{zhang2017age}
Z.~Zhang, Y.~Song, and H.~Qi.
\newblock Age progression/regression by conditional adversarial autoencoder.
\newblock In {\em IEEE Conference on Computer Vision and Pattern Recognition
  (CVPR)}, 2017.

\bibitem{zheng2018crossnet}
H.~Zheng, M.~Ji, H.~Wang, Y.~Liu, and L.~Fang.
\newblock {CrossNet}: An end-to-end reference-based super resolution network
  using cross-scale warping.
\newblock In {\em European Conference on Computer Vision (ECCV)}, 2018.

\bibitem{zhu2014single}
Y.~Zhu, Y.~Zhang, and A.~L. Yuille.
\newblock Single image super-resolution using deformable patches.
\newblock In {\em IEEE Conference on Computer Vision and Pattern Recognition
  (CVPR)}, 2014.

\end{thebibliography}
}

\end{document}